\documentclass{article}
\usepackage{PRIMEarxiv}
\usepackage[T1]{fontenc}
\usepackage[utf8]{inputenc}
\usepackage{amsmath,amssymb,amsthm}
\usepackage{array}
\usepackage{booktabs}
\usepackage{enumitem}
\usepackage{graphicx}
\usepackage{float}
\usepackage{hyperref}
\usepackage{lineno}
\usepackage{microtype}
\usepackage{multirow}
\usepackage[most]{tcolorbox}
\usepackage{url}
\usepackage{xcolor}
\usepackage{colortbl}

\IfFileExists{algorithm.sty}{
  \usepackage{algorithm}
}{
  \usepackage{float}
  \newfloat{algorithm}{tbp}{loa}
  \floatname{algorithm}{Algorithm}
}
\IfFileExists{algpseudocode.sty}{
  \usepackage{algpseudocode}
}{
  \newcounter{ALGline}
  \newenvironment{algorithmic}[1][]{
    \setcounter{ALGline}{0}
    \begin{list}{}{
      \setlength{\leftmargin}{2.2em}\setlength{\labelwidth}{1.8em}
      \setlength{\labelsep}{0.4em}\setlength{\itemsep}{0.15ex}
      \setlength{\topsep}{0.25ex}\setlength{\parsep}{0pt}\setlength{\partopsep}{0pt}
    }
  }{
    \end{list}
  }
  \newcommand{\State}{\stepcounter{ALGline}\item[\arabic{ALGline}.]}
  \newcommand{\Require}{\State \textbf{Require:} }
  
  \newcommand{\Return}{\textbf{return} }
  
}

\newtcblisting{promptbox}[1][]{
  colback=gray!5, colframe=gray!60!black, fonttitle=\bfseries,
  colbacktitle=gray!85!black, title=#1, breakable, listing only,
  listing options={basicstyle=\ttfamily\small, breaklines=true, columns=fullflexible}
}

\newtcblisting{examplebox}[1][]{
  colback=paperteallight, colframe=paperteal, fonttitle=\bfseries,
  colbacktitle=paperteal, title={#1}, breakable, listing only,
  listing options={basicstyle=\ttfamily\footnotesize, breaklines=true, columns=fullflexible}
}

\definecolor{darkblue}{rgb}{0, 0, 0.5}
\definecolor{paperteal}{RGB}{48,121,137}
\definecolor{paperteallight}{RGB}{235,244,246}
\definecolor{papergraylight}{RGB}{244,244,244}

\newcommand{\graph}{\mathcal{G}}        
\newcommand{\ledger}{\mathcal{L}}       

\title{AutoMem: A Text-Gradient Recursive Self-Improvement Framework for Automated Memory Architectures Search}

\author{
  Lin Du$^{1}$, Jie Zhou$^{1,2}$\footnotemark[1], Yuxuan Cai$^{1}$, Kai Chen$^{2}$, Qin Chen$^1$, Xin Li$^{2}$, Bo Zhang$^{2}$, Wei Li$^{2}$, Liang He$^1$ \\
  $^1$ School of Computer Science and Technology, East China Normal University, Shanghai\\ 
  $^2$ Shanghai AI Laboratory \\
  \texttt{\{jzhou, qchen, lhe\}@cs.ecnu.edu.cn} \\
  \textcolor{red}{\url{https://github.com/ECNU-ICALK/AutoMem}} 
}

\begin{document}

\maketitle
\footnotetext[1]{Corresponding Author}
\begin{abstract}
Long-term memory is increasingly central to LLM agents, yet memory design remains a highly coupled architecture problem: what to encode, how to store it, how to retrieve it, and how to manage it can vary substantially across tasks and backbone models. We construct a discrete search space with 5 encoders, 5 stores, 6 retrievers, and 4 managers, and show that no single memory architecture consistently dominates: different tasks favor different module combinations, leading to substantial performance gaps. Motivated by this, we propose \textsc{AutoMem}, a text-gradient recursive self-improvement framework for task-adaptive memory architecture search. \textsc{AutoMem} optimizes over the factored space through two components: Experience-Guided Architecture Search, which proposes candidate architectures from historical search trajectories and accumulated reflections, and Failure-Guided Module Diagnosis, which localizes memory-related failures to specific modules and converts them into targeted textual feedback. Experiments on GAIA, WebWalkerQA, and xBench-DeepSearch across two LLM backbones show that \textsc{AutoMem} consistently discovers task-adaptive memory architectures that outperform the strongest human-designed memory baselines, improving accuracy by $2.8$ points on average across six benchmark-backbone settings. Further analysis shows that \textsc{AutoMem} achieves a favorable accuracy-efficiency trade-off, reducing token cost by $14.3\%$ over the strongest accuracy baselines under Qwen3.5-122B-A10B, while also finding stronger architectures than substantially larger random searches within only a few guided iterations.
\end{abstract}

\section{Introduction}
\label{sec:intro}
Long-term memory has become an important component of LLM agents \cite{cai2025building}. It enables agents to preserve reusable information from past interactions, trajectories, and task executions, and to reuse such information in future decision making. Existing work has explored episodic memory over dialogues and trajectories \cite{park2023generative,packer2024memgpt,chhikara2025mem0,xu2025amem}, verbal reflection and distilled experience \cite{shinn2023reflexion,zhao2024expel}, procedural memory in the form of skills, workflows, or test-time cheatsheets \cite{wang2023voyager,wang2024awm,zheng2025skillweaver,suzgun2025dynamiccheatsheet}, and structured experience sharing across agents \cite{tang2025agentkb}. These systems show that agent memory is no longer a single retrieval-augmented store, but a multi-module subsystem involving decisions about what to encode, how to store it, how to retrieve it, and how to maintain it over time \cite{du2026memorysurvey}.

However, most memory systems are still manually designed and fixed once deployed. This is limiting because memory effectiveness depends on interactions among multiple modules. A task may fail because useful information was not encoded, because the stored representation is hard to retrieve, because the retriever selects irrelevant memories, or because the memory pool contains stale, duplicated, or conflicting entries. Therefore, improving agent memory requires optimizing the full memory architecture rather than tuning a single retrieval component.

To study this problem systematically, we factor long-term memory into four modules: Encode, Store, Retrieve, and Manage. Encode decides what information should be written into memory, Store decides how the memory is represented, Retrieve decides how relevant memories are selected, and Manage controls memory consolidation, deduplication, eviction, and conflict resolution. Based on common designs in prior memory systems, we construct a discrete search space with 5 encoders, 5 stores, 6 retrievers, and 4 managers. This turns memory design into a memory architecture search problem over $\mathcal{A}=\mathcal{E}\times\mathcal{S}\times\mathcal{R}\times\mathcal{M}$.

Our pilot study shows that architecture choice is a major source of performance variation. Simple search over the factored space can discover architectures that match or outperform strong fixed memory baselines. More importantly, the best architecture differs across datasets and backbone models, and module effects are strongly coupled. For example, a strong retriever may still fail when the encoder writes low-quality memories, while a useful store format may be ineffective if the retriever cannot query it properly. These findings suggest that memory architecture search is valuable, but exhaustive or random search is expensive because each candidate requires a full agent rollout and provides little reusable optimization signal.

We propose \textsc{AutoMem}, a text-gradient recursive self-improvement framework for task-adaptive memory architecture search (Figure~\ref{fig:framework}). \textsc{AutoMem} contains two components. Experience-Guided Architecture Search proposes candidate architectures from historical search trajectories, evaluated architectures, performance records, and accumulated reflections. Failure-Guided Module Diagnosis analyzes failed rollouts, localizes memory-related failures to Encode, Store, Retrieve, or Manage, and converts them into targeted textual feedback for the next search round. Through repeated proposal, evaluation, diagnosis, and validation, \textsc{AutoMem} turns failed trajectories into directional signals for improving memory architectures. We evaluate \textsc{AutoMem} on GAIA, WebWalkerQA, and xBench-DeepSearch across multiple LLM backbones. Experiments show that \textsc{AutoMem} discovers task-adaptive memory architectures that outperforms all the strong baselines, while reducing per-task token cost. Moreover, \textsc{AutoMem} steadily improves the quality of discovered memory architectures across iterations and quickly identifies superior architectures within only a few rounds, outperforming those found by substantially larger random searches.

Our contributions are threefold:
\begin{itemize}[leftmargin=1.2em, itemsep=0.15em, topsep=0.2em]
\item We systematize the design space of long-term memory for LLM agents by factorizing memory architectures into Encode, Store, Retrieve, and Manage modules, and empirically study which module designs and cross-module interactions are most critical to memory effectiveness.
\item We propose \textsc{AutoMem}, a recursive self-improvement framework for efficiently identifying task-adaptive memory architectures under limited evaluation budgets, combining experience-guided architecture search with failure-guided module diagnosis.
\item We conduct experiments on GAIA, WebWalkerQA, and xBench-DeepSearch across multiple LLM backbones, showing that task-adaptive memory architectures outperform human-design framework.
\end{itemize}

\section{Related Work}
\label{sec:related}

\paragraph{Memory architectures for LLM agents.}
Long-term memory has become an important component of LLM agents. Existing methods store and retrieve episodic experience by summarizing dialogues or trajectories into memory stores and retrieving them by relevance \cite{park2023generative,packer2024memgpt,chhikara2025mem0,xu2025amem}, distill verbal self-reflections and cross-task insights for reuse \cite{shinn2023reflexion,zhao2024expel}, induce procedural memory such as reusable skills, workflows, or test-time ``cheatsheets'' \cite{wang2023voyager,wang2024awm,zheng2025skillweaver,suzgun2025dynamiccheatsheet}, or share structured experience across agents \cite{tang2025agentkb}. These studies demonstrate that memory can improve agent learning, reasoning, and reuse across tasks. However, they also show that memory is not a single retrieval component, but a coupled architecture involving what to encode, how to store it, how to retrieve it, and how to maintain it over time \cite{du2026memorysurvey}. Most existing systems manually choose one memory configuration and keep it fixed. In contrast, \textsc{AutoMem} systematizes these design choices into a discrete Encode/Store/Retrieve/Manage space and treats the memory architecture itself as an optimizable variable.

\paragraph{Self-Evolving Memory.}
A closer line of work studies adaptive or self-evolving memory systems. Zhang et al. \cite{zhang2025memevolve} meta-evolves the whole memory system by synthesizing new memory implementations from execution logs and selecting candidates through tournament evaluation. This approach increases the flexibility of memory design, but each candidate is a monolithic implementation, making it difficult to assign credit to individual memory modules or perform controlled module-level edits. Liu et al. \cite{liu2026evolvemem} introduces a structured diagnosis loop with revert-on-regression, but it mainly adapts retrieval configurations for dialogue-QA memory. More broadly, automated agent design methods search over code-defined agents \cite{hu2025adas}, workflows \cite{zhang2025aflow}, agent graphs \cite{zhuge2024gptswarm}, symbolic agent parameters \cite{zhou2024symbolic}, or self-modifying coding agents \cite{zhang2026dgm}, but they mainly optimize prompts, tools, workflows, or orchestration rather than the memory subsystem. \textsc{AutoMem} differs from these methods by performing structured, per-module-attributable search over all four memory modules. This enables valid-by-construction candidates, controlled architecture edits, and task-adaptive memory optimization on agentic web tasks.

\paragraph{Text-gradient Optimization.}
\textsc{AutoMem} is also related to textual-gradient and failure-driven optimization. Prior work treats LLM-generated natural-language feedback as an optimization signal. \textsc{ProTeGi} uses textual ``gradients'' to critique and edit prompts \cite{pryzant2023protegi}, \textsc{TextGrad} propagates textual feedback through a computation graph \cite{yuksekgonul2024textgrad}, and OPRO and \textsc{DSPy} use LLMs to optimize prompts or pipeline parameters \cite{yang2024opro,khattab2023dspy}. These methods mainly optimize a single text object, prompt, or pipeline parameter, rather than a structured memory architecture. Another line attributes agent failures to responsible steps or modules \cite{zhang2025whoandwhen,zhang2025agentracer,zhu2025agentdebug}, but usually stops at diagnosis. \textsc{AutoMem} connects these two directions by converting failed rollouts into module-level textual feedback. The feedback localizes memory-related failures to Encode, Store, Retrieve, or Manage, and then guides architecture edits that are validated across search rounds. This closes the loop from failure attribution to memory architecture improvement.

\begin{table}[t]
  \centering
  \caption{Representative memory systems decomposed along the four modules
  $(E,S,R,M)$, each mapped to the nearest atomic component in our menu
  (Table~\ref{tab:component-lineage}); the Manage column instead lists each
  system's lifecycle operations, which the Manage presets in that menu bundle.
  ``--'' marks a module a system does
  not implement (e.g.\ no selective retrieval, or no lifecycle management). Prior
  systems instantiate only a few components and leave the rest at generic
  defaults, whereas \textsc{AutoMem} searches all four.}
  \label{tab:prior-decomposition}
  \small\setlength{\tabcolsep}{4pt}
  \resizebox{\textwidth}{!}{%
  \begin{tabular}{l llll}
    \toprule
    System & Encode & Store & Retrieve & Manage \\
    \midrule
    Generative Agents \cite{park2023generative}           & \texttt{trajectory}, \texttt{insight} & \texttt{vector} & \texttt{cbr\_rerank} & \texttt{reflect}, \texttt{merge} \\
    MemGPT \cite{packer2024memgpt}                        & \texttt{trajectory}                   & \texttt{vector} & \texttt{hybrid}      & \texttt{forget} \\
    Mem0 \cite{chhikara2025mem0}                          & \texttt{tip}                          & \texttt{vector} & \texttt{hybrid}      & \texttt{update}, \texttt{delete} \\
    A-MEM \cite{xu2025amem}                               & \texttt{trajectory}                   & \texttt{graph}     & \texttt{graph}       & \texttt{merge}, \texttt{evolve} \\
    Zep \cite{rasmussen2025zep}                           & \texttt{trajectory}                   & \texttt{llm\_graph} & \texttt{graph}       & \texttt{merge}, \texttt{forget} \\
    MemoryBank \cite{zhong2023memorybank}                 & \texttt{trajectory}                   & \texttt{vector} & \texttt{hybrid}      & \texttt{forget} \\
    Reflexion \cite{shinn2023reflexion}                   & \texttt{insight}                      & \texttt{json}   & --                   & -- \\
    ExpeL \cite{zhao2024expel}                            & \texttt{tip}, \texttt{insight}        & \texttt{hybrid} & \texttt{contrastive} & \texttt{update}, \texttt{merge} \\
    Voyager \cite{wang2023voyager}                        & \texttt{shortcut}                     & \texttt{vector} & \texttt{hybrid}      & \texttt{validate} \\
    Agent Workflow Memory \cite{wang2024awm}              & \texttt{workflow}                     & \texttt{json}   & --                   & \texttt{merge} \\
    Agent-KB \cite{tang2025agentkb}                       & \texttt{tip}, \texttt{workflow}       & \texttt{hybrid} & \texttt{cbr\_rerank} & \texttt{merge}, \texttt{validate} \\
    Dynamic Cheatsheet \cite{suzgun2025dynamiccheatsheet} & \texttt{tip}, \texttt{shortcut}       & \texttt{json}   & --                   & \texttt{update}, \texttt{validate} \\
    Memp \cite{fang2025memp}                              & \texttt{trajectory}, \texttt{workflow} & \texttt{vector} & \texttt{hybrid}     & \texttt{update}, \texttt{merge} \\
    \bottomrule
  \end{tabular}%
  }
\end{table}

\section{Preliminary Analysis}
\label{sec:preliminary}

\subsection{A Modular View of Agent Memory Architectures}
\label{sec:prelim-memory-space}

Before introducing \textsc{AutoMem}, we first clarify what is being optimized when an LLM agent is equipped with long-term memory. Existing memory systems differ substantially in surface form: some store episodic trajectories, some distill verbal reflections, some maintain skill libraries or workflow memories, and others construct graph-structured memories or curated knowledge bases. Despite this diversity, their designs can be organized into four recurring architectural modules: \textbf{Encode}, \textbf{Store}, \textbf{Retrieve}, and \textbf{Manage}. We use $\mathcal{E}$, $\mathcal{S}$, $\mathcal{R}$, and $\mathcal{M}$ to denote the candidate sets of these four modules, and use lowercase $e \in \mathcal{E}$, $s \in \mathcal{S}$, $r \in \mathcal{R}$, and $m \in \mathcal{M}$ to denote concrete module choices.

\textbf{Encode} determines what information should be written into memory. Common choices include task-agnostic tips, failure-derived insights, compressed action--observation trajectories, reusable workflows, and parameterized skills or shortcuts. These choices reflect different assumptions about what kind of experience is reusable: natural-language reflections emphasize general lessons, trajectory memories preserve concrete demonstrations, while workflow and skill memories aim to capture procedural knowledge.

\textbf{Store} determines how encoded memories are represented. Prior systems use plain textual buffers, key--value records, dense vector indices, hybrid symbolic--vector stores, entity--relation graphs, or LLM-constructed temporal knowledge graphs. The store format affects not only memory capacity, but also which retrieval and management operations are feasible. For example, graph-based retrieval requires a graph-structured store, while embedding-based retrieval is naturally paired with vectorized memory representations.

\textbf{Retrieve} determines how relevant memories are selected at inference time. Existing approaches include semantic retrieval, lexical--semantic hybrid retrieval, contrastive retrieval over success and failure cases, case-based reranking, graph traversal, hypothetical-document embeddings, and diversity-aware selection. Retrieval is often treated as the central memory operation, but its effectiveness depends heavily on what has been encoded and how it has been stored.

\textbf{Manage} determines how the memory pool evolves over time. Representative operations include forgetting, deduplication, conflict resolution, memory consolidation, trajectory-to-workflow promotion, and skill validation. Management is especially important for long-running agents, where stale, redundant, or conflicting memories can degrade performance even when retrieval itself is accurate.

Under this modular view, a memory architecture is a tuple $a=(e,s,r,m)$, where $e$, $s$, $r$, and $m$ are selected from the Encode, Store, Retrieve, and Manage candidate sets, respectively. The unconstrained Cartesian space is $\widetilde{\mathcal{A}}=\mathcal{E}\times\mathcal{S}\times\mathcal{R}\times\mathcal{M}$. In practice, not every tuple is valid because some modules impose compatibility constraints. We therefore define the feasible memory architecture space as
\[
\mathcal{A}=\{(e,s,r,m)\in\widetilde{\mathcal{A}}\mid \operatorname{Valid}(e,s,r,m)=1\},
\]
where $\operatorname{Valid}(\cdot)$ filters out incompatible module combinations, such as pairing a graph-based retriever with a non-graph store. This definition separates module candidate sets from concrete memory architectures and ensures that every searched architecture is valid by construction.

This four-module formulation reveals that existing memory systems are not incomparable monoliths, but fixed points in a shared architectural space. Table~\ref{tab:prior-decomposition} decomposes representative memory systems along Encode, Store, Retrieve, and Manage. Two patterns emerge. First, prior systems typically specialize in only one or two modules. For example, skill-library and workflow-memory agents invest heavily in \textbf{Encode}, graph-memory systems emphasize \textbf{Store}, while memory banks and self-updating memory systems focus more on \textbf{Manage}. Second, because most systems keep the remaining modules at generic defaults, they are difficult to adapt when failures originate outside their specialized component. A failure caused by poor encoding cannot be solved by a stronger retriever alone; similarly, a useful memory representation may remain ineffective if retrieval or lifecycle management is mismatched.

These observations motivate a shift from designing one fixed memory system to searching over memory architectures. Rather than asking which existing memory method is universally best, we ask which feasible combination $a=(e,s,r,m)\in\mathcal{A}$ is most suitable for a given task distribution and backbone model.

\subsection{Pilot Study: Why Memory Architecture Search Is Necessary}
\label{sec:prelim-pilot}

We conduct a preliminary random-search study to examine whether the feasible architecture space $\mathcal{A}$ contains useful task-specific memory architectures and whether blind search is sufficient to find them efficiently. In this probe, we uniformly sample valid architectures $a\sim\operatorname{Unif}(\mathcal{A})$ after filtering out incompatible module combinations. Each sampled architecture is instantiated as an independent memory system and evaluated on the same task batches. The score of one trial does not affect the next trial; no failure attribution, module-level editing, or historical search memory is used.

\begin{figure}[t]
  \centering
  \begin{minipage}{0.49\linewidth}
    \centering
    \includegraphics[width=\linewidth]{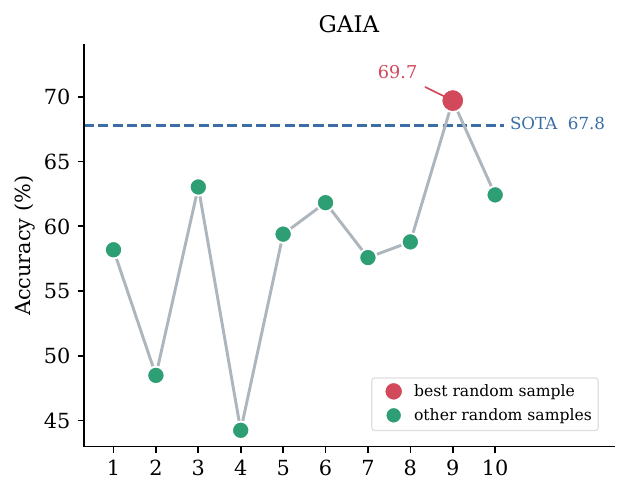}
  \end{minipage}\hfill
  \begin{minipage}{0.49\linewidth}
    \centering
    \includegraphics[width=\linewidth]{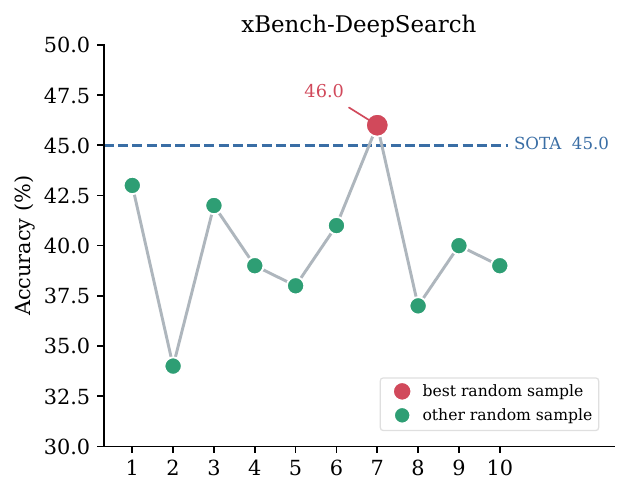}
  \end{minipage}
  \caption{\textbf{Strong but scattered memory architectures exist on both
  benchmarks.} Each marker is one memory architecture sampled from the factored
  space, re-evaluated on the full benchmark (Qwen3.5-122B-A10B; horizontal axis is
  a layout index). On GAIA (left) and xBench-DeepSearch (right), the best sample
  (red) exceeds the SOTA reference (dashed line): $69.7$ vs.\ $67.8$ and $46.0$
  vs.\ $45.0$. The reference is the strongest same-backbone fixed memory baseline
  (MemoryBank) on GAIA and the published SOTA on xBench-DeepSearch.}
  \label{fig:random-gold}
\end{figure}

\paragraph{Finding 1: Strong memory architectures exist, but they are task-specific.} The random-search probe shows that $\mathcal{A}$ contains architectures that can outperform strong fixed memory baselines. For example, the best sampled architecture reaches $69.7\%$ on GAIA, improving over the strongest fixed memory baseline under the same backbone ($67.8\%$, MemoryBank). This suggests that the architecture space is worth optimizing: fixed memory designs do not necessarily represent the ceiling of memory-enhanced agent performance. The same holds beyond GAIA: on xBench-DeepSearch the best sampled architecture reaches $46.0\%$, above the SOTA reference of $45.0\%$. Figure~\ref{fig:random-gold} plots both benchmarks, showing that strong architectures exist on each, yet appear as scattered, dataset-dependent peaks.

However, the best architectures are not universal. Across GAIA, WebWalkerQA, and xBench-DeepSearch, the top configurations differ across all four modules. For instance, one task may favor workflow-style encoding with a hybrid store and semantic retrieval, whereas another may favor shortcut-style encoding with a graph store and contrastive retrieval. Moreover, a memory system that performs strongly on one benchmark can underperform the no-memory baseline on another. These results indicate that memory effectiveness depends on the interaction among task distribution, backbone model, and module composition.

\paragraph{Finding 2: Module effects are coupled rather than additive.} The pilot study also shows that individual module choices cannot be evaluated in isolation. The contribution of an encoder $e$ depends on whether the store $s$ can preserve its outputs in a retrievable form; the value of a retriever $r$ depends on whether the encoded memories are informative and whether the store supports the required retrieval operation; and the effect of a manager $m$ depends on the density, redundancy, and stability of the memory pool produced by the other modules. A strong retriever may fail when the encoder writes low-quality memories, a graph store may be ineffective without a compatible retrieval policy, and aggressive memory management can remove useful information when the encoded memory is sparse. Thus, memory architecture quality is determined by cross-module compatibility rather than by the independent quality of each module.

\paragraph{Finding 3: Random search obtains SOTA results but wastes evaluation budget.} Although random search can occasionally discover strong architectures, it cannot turn failed trials into reusable search signal. Each candidate $a$ requires a costly agent rollout, yet the next candidate is sampled independently of previous successes and failures. Increasing the sampling budget only increases the chance of accidentally landing on a good configuration; it does not make the search process more informed. This limitation is especially problematic because the search landscape is scattered and task-dependent, with strong architectures appearing as isolated peaks rather than as a smooth region.

The preliminary analysis shows that strong memory architectures exist, but their effectiveness is task-specific, module-coupled, and difficult to discover through blind search. This motivates a formal task definition: selecting a feasible memory architecture $a\in\mathcal{A}$ that best matches a given task distribution, backbone model, and evaluation budget.

\subsection{Definition: Task-Adaptive Memory Architecture Selection}
\label{sec:prelim-formulation}

Based on the modular view above, we formulate memory design as a task-adaptive architecture selection problem. Let $\mathcal{D}$ denote a task distribution and let $B=\{(x_i,y_i)\}_{i=1}^{n}\sim\mathcal{D}$ denote an evaluation batch sampled from this distribution. Let $A_a$ denote the backbone agent equipped with memory architecture $a\in\mathcal{A}$, and let $A_0$ denote the same backbone agent without long-term memory. For a task input $x_i$, the prediction made by $A_a$ is denoted as $\hat{y}_i^{a}=A_a(x_i)$. We evaluate a candidate architecture by its task accuracy:
\[
\mathrm{Acc}(A_a;B)=\frac{1}{|B|}\sum_{i=1}^{|B|}\mathbf{1}\left[\hat{y}_i^{a}=y_i\right].
\]

In addition to absolute accuracy, we consider whether the selected architecture actually improves the backbone agent rather than merely inheriting its original capability. We therefore define the memory-induced gain of architecture $a$ on batch $B$ as
\[
\Delta(a;B)=\mathrm{Acc}(A_a;B)-\mathrm{Acc}(A_0;B).
\]
A desirable memory architecture should achieve high $\mathrm{Acc}(A_a;B)$, provide positive and stable $\Delta(a;B)$, and avoid unnecessary inference cost. When two architectures obtain comparable accuracy, we prefer the one with lower rollout cost, such as lower token consumption or fewer memory operations.

Given a search budget $T$, the goal is to identify an architecture
\[
a^\star \in \arg\max_{a\in\mathcal{A}} \ \mathbb{E}_{B\sim\mathcal{D}}\left[\mathrm{Acc}(A_a;B)\right],
\quad \text{s.t.} \quad N_{\mathrm{eval}}\le T,
\]
where $N_{\mathrm{eval}}$ is the number of evaluated candidate architectures. In practice, the expectation over $\mathcal{D}$ is estimated using held-out task batches, and we report both task accuracy and memory-induced gains over $A_0$.

This formulation clarifies the scope of our optimization. We do not update the backbone model parameters, and we do not synthesize arbitrary memory implementations. Instead, we search over a structured and feasible space $\mathcal{A}$ of memory architectures and select the combination of encoding, storage, retrieval, and management modules that best fits a given task distribution. Since evaluating each candidate requires running the full agent, exhaustive search is expensive and random search provides limited reusable signal. This motivates \textsc{AutoMem}, which uses previous trials and failure analyses to guide memory architecture search under a limited evaluation budget.
\section{Method}
\label{sec:method}

\begin{figure}[t]
  \centering
  \includegraphics[width=\linewidth]{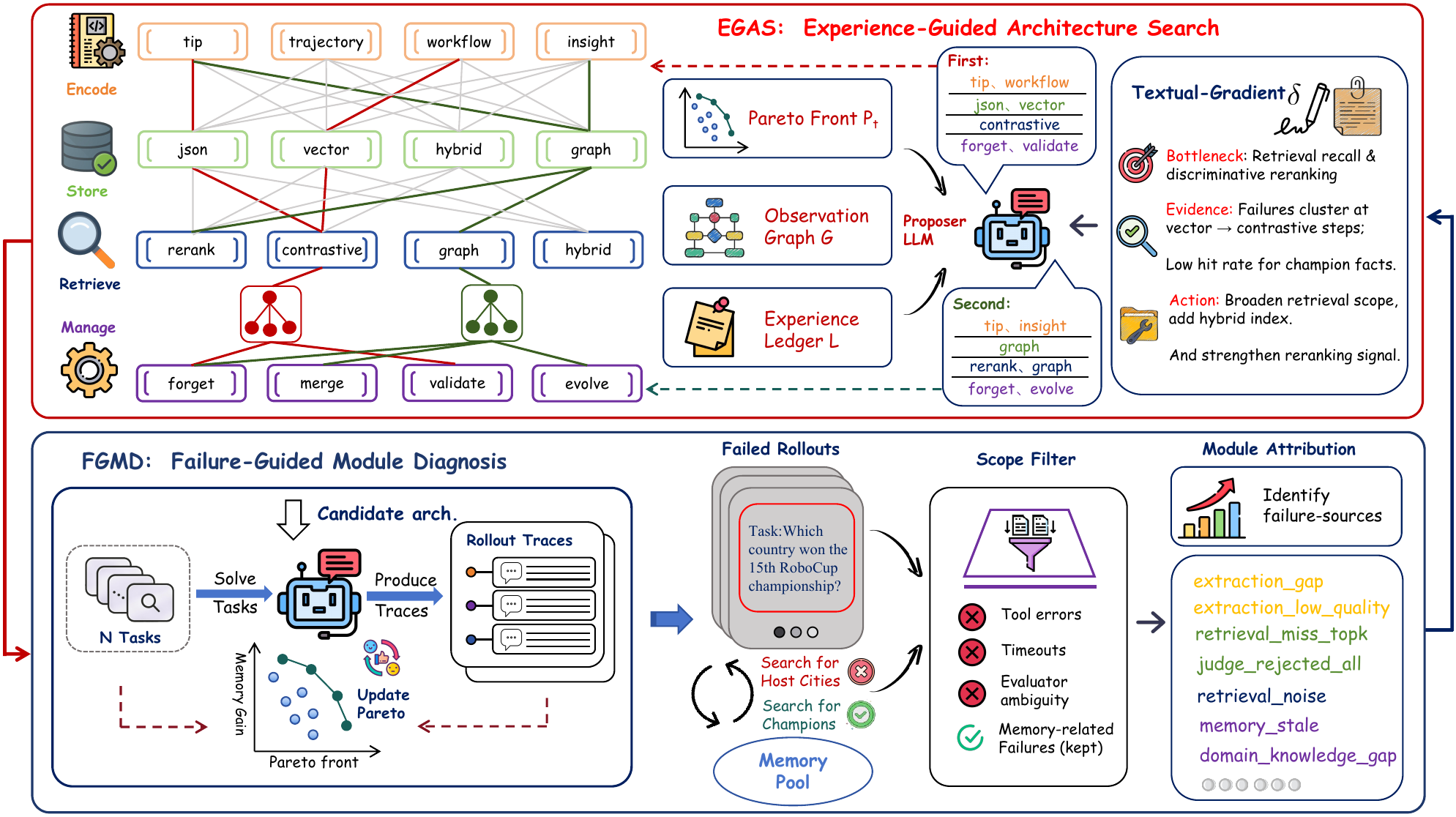}
  \caption{\textbf{Overview of \textsc{AutoMem}.} \textsc{AutoMem} performs
  text-gradient search over the factored memory architecture space
  $\mathcal{A}=\mathcal{E}\times\mathcal{S}\times\mathcal{R}\times\mathcal{M}$
  (left lattice: Encode, Store, Retrieve, and Manage options; the two
  highlighted paths are two candidate architectures).
  \textbf{Top---Experience-Guided Architecture Search (EGAS,
  \S\ref{sec:method-egas}):} a proposer LLM conditions on the Pareto front
  $P_t$, the observation graph $\graph$, the experience ledger $\ledger$, and
  the latest textual gradient $\delta_t$ to propose $K$ feasible candidate
  architectures.
  \textbf{Bottom---Failure-Guided Module Diagnosis (FGMD,
  \S\ref{sec:method-fgmd}):} candidates are evaluated on the search batch
  $B_s$, producing rollout traces and updating the Pareto front; a rule-based
  scope gate discards memory-unrelated failures (e.g., tool errors, timeouts),
  and the remaining failures are attributed to the responsible module, yielding
  the textual gradient $\delta_{t+1}$ that steers the next round.}
  \label{fig:framework}
\end{figure}

\subsection{Overview}
\label{sec:method-overview}

We propose \textsc{AutoMem}, a text-gradient recursive self-improvement framework
for the task-adaptive memory architecture selection problem of
\S\ref{sec:prelim-formulation}: within an evaluation budget $T$, find the feasible
architecture $a=(e,s,r,m)\in\mathcal{A}$ (\S\ref{sec:prelim-memory-space}) that
maximizes task accuracy $\mathrm{Acc}(A_a;B)$ while providing positive
memory-induced gain $\Delta(a;B)$ over the no-memory agent $A_0$.
\textsc{AutoMem} instantiates the expectation over $\mathcal{D}$ with three
disjoint batches---a search batch $B_s$ on which candidates are evaluated, a
validation batch $B_v$ on which the current best architecture must confirm its
gain before being accepted, and a held-out test batch $B_t$ for final
reporting---and ranks candidates by a Pareto objective over task accuracy,
memory-induced gain, and rollout cost.

The key challenge is that $\mathcal{A}$ is discrete, module-coupled, and observable
only through full agent rollouts. Therefore, \textsc{AutoMem} does not rely on
analytic gradients. Instead, it constructs a textual optimization signal from failed
rollouts. Each round contains two components, as illustrated in
Figure~\ref{fig:framework}. First,
Experience-Guided Architecture Search (EGAS) proposes candidate
architectures by conditioning a proposer LLM on historical search experience,
current Pareto-optimal architectures, and the latest diagnostic feedback. Second,
Failure-Guided Module Diagnosis (FGMD) analyzes unsuccessful rollouts,
localizes failures to one of the four memory modules, and converts the evidence into
module-level textual feedback for the next round.

Concretely, \textsc{AutoMem} first evaluates the no-memory baseline and initializes
two search memories: an experience ledger $\mathcal{L}$ and an observation graph
$\mathcal{G}$. At round $t$, EGAS proposes $K$ feasible architectures
$\{a_t^{(k)}\}_{k=1}^{K}$. Each candidate is evaluated on $B_s$, producing task
accuracy, memory-induced gain, cost, and rollout traces. The Pareto front $P_t$ is
then updated. FGMD analyzes failed traces from the round and produces a textual
gradient $\delta_{t+1}$, which summarizes the dominant module bottleneck and
recommends targeted architecture edits. The loop repeats for $N$ rounds, and the
final architecture is selected by validation performance on $B_v$ and reported on
the held-out test batch $B_t$. Algorithm~\ref{alg:automem} summarizes the overall
procedure, and Appendix~\ref{app:worked} traces one optimization round of this
loop with verbatim runtime artifacts.

\begin{algorithm}[t]
\caption{\textsc{AutoMem}: Text-gradient search over memory architectures.}
\label{alg:automem}
\begin{algorithmic}[1]
\Require Feasible architecture space $\mathcal{A}$; search, validation, and test
batches $B_s,B_v,B_t$; rounds $N$; candidates per round $K$; validation period $\tau$.
\State Evaluate the no-memory agent $A_0$ on $B_s\cup B_v$.
\State Initialize experience ledger $\mathcal{L}$, observation graph $\mathcal{G}$,
Pareto front $P_0\leftarrow\emptyset$, and textual feedback $\delta_1\leftarrow\emptyset$.
\For{$t=1$ \textbf{to} $N$}
    \State $\{a_t^{(1)},\ldots,a_t^{(K)}\}
    \leftarrow
    \textsc{EGAS}(\delta_t,\mathcal{L},\mathcal{G},P_{t-1})$.
    \For{$k=1$ \textbf{to} $K$ \textbf{in parallel}}
        \State Run agent $A_{a_t^{(k)}}$ on $B_s$ and record accuracy, gain, cost,
        and rollout traces.
    \EndFor
    \State Update $P_t$ using non-dominated candidates under accuracy, gain, and cost.
    \If{$t \bmod \tau = 0$}
        \State Validate the current Pareto-front leader on $B_v$.
    \EndIf
    \State $\delta_{t+1}
    \leftarrow
    \textsc{FGMD}(\{\text{failed rollouts of }a_t^{(k)}\}_{k=1}^{K})$.
    \State Update $\mathcal{L}$ and $\mathcal{G}$ with round-level outcomes and
    module-level evidence.
\EndFor
\State \Return the validated Pareto-front leader $a^\star$, evaluated on $B_t$.
\end{algorithmic}
\end{algorithm}

\subsection{Experience-Guided Architecture Search}
\label{sec:method-egas}

Experience-Guided Architecture Search proposes feasible memory architectures by
using accumulated search evidence as a structured prior. A naive proposer that only
uses the latest diagnostic feedback may repeatedly explore already failed edits or
overfit to a small set of recent trajectories. EGAS addresses this by conditioning
candidate generation on three forms of search state: the Pareto front $P_t$, the
experience ledger $\mathcal{L}$, and the observation graph $\mathcal{G}$.

\paragraph{Candidate proposal.}
At round $t$, EGAS receives the latest textual feedback $\delta_t$ and proposes
$K$ architectures:
\[
\{a_t^{(1)},\ldots,a_t^{(K)}\}
=
\mathrm{Propose}_{\theta}
\left(
\delta_t,\mathcal{L}_t,\mathcal{G}_t,P_t
\right),
\]
where $\mathrm{Propose}_{\theta}$ is an LLM-based proposer. Each proposal is a tuple
$a=(e,s,r,m)$ rather than a regenerated memory implementation. This design keeps the
search controlled: one proposal changes named coordinates in the architecture space
while preserving validity constraints. We enforce validity with a deterministic
checker:
\[
a_t^{(k)}\in\mathcal{A}
\quad\Longleftrightarrow\quad
\operatorname{Valid}(e_t^{(k)},s_t^{(k)},r_t^{(k)},m_t^{(k)})=1.
\]
Invalid proposals are rejected and resampled before evaluation. A verbatim
proposal, together with the diagnostic evidence it cites, is shown in
Appendix~\ref{app:worked-round}.

\paragraph{Experience ledger.}
The ledger $\mathcal{L}$ stores reusable cross-round search knowledge. Each entry is
represented as
\[
\ell =
(c,\ u,\ q,\ z),
\]
where $c$ is the condition under which the entry applies, $u$ is a recommended or
discouraged edit, $q$ is supporting evidence, and $z$ is the status of the entry
(\emph{active}, \emph{pending}, or \emph{refuted}). For example, if repeated trials
show that graph retrieval underperforms when the memory pool is sparse, the ledger
records this as a discouraged Retrieve--Store edit under sparse-memory conditions.
During proposal, active entries bias the LLM toward edits with positive evidence,
while refuted or dead-end entries prevent repeated exploration of previously
unproductive directions. Appendix~\ref{app:worked-ledger} shows verbatim ledger
entries, including a principle refuted by differential validation.

\paragraph{Observation graph.}
The observation graph $\mathcal{G}$ stores task-pattern-level statistics about
which module choices have worked in previous rounds. Nodes correspond to task
patterns, module choices, and observed outcomes; edges record empirical associations
between them. For a task pattern $c$ and a module choice $u$, EGAS maintains running
statistics such as
\[
\mu(c,u)
=
\frac{1}{n(c,u)}
\sum_{j:\,c_j=c,\,u_j=u}
\Delta(a_j;B_j),
\]
where $n(c,u)$ is the number of evaluated architectures containing choice $u$ under
condition $c$. These statistics are used as soft priors rather than hard rules. This
is especially useful for the Encode module, since encoding failures are often
under-observed: a missing memory is harder to attribute than an incorrectly
retrieved one. An excerpt of $\mathcal{G}$ is shown in
Appendix~\ref{app:worked-graph}.

\paragraph{Pareto-based acceptance.}
After evaluating all proposed candidates, EGAS updates a Pareto front over accuracy,
memory-induced gain, and cost. A candidate $a_i$ dominates $a_j$ if it is no worse
on all objectives and strictly better on at least one:
\[
a_i \succ a_j
\quad\Longleftrightarrow\quad
\mathrm{Acc}_i\ge \mathrm{Acc}_j,\ 
\Delta_i\ge \Delta_j,\ 
\mathrm{Cost}_i\le \mathrm{Cost}_j,
\]
with at least one strict inequality. The updated front is
\[
P_{t+1}
=
\left\{
a\in P_t\cup\{a_t^{(k)}\}_{k=1}^{K}
\mid
\nexists a'\in P_t\cup\{a_t^{(k)}\}_{k=1}^{K}
\text{ such that } a'\succ a
\right\}.
\]
The front leader is periodically re-evaluated on validation data to reduce search
batch overfitting. Only validated improvements are promoted as reliable experience
in $\mathcal{L}$ and $\mathcal{G}$.

\subsection{Failure-Guided Module Diagnosis}
\label{sec:method-fgmd}

Failure-Guided Module Diagnosis converts failed rollouts into a module-level textual
gradient. Given failed trajectories from the current round, FGMD answers three
questions: whether the failure is memory-attributable, which module is most likely
responsible, and what architecture edit should be attempted next.

\paragraph{Scope filtering.}
Not every failed rollout should update the memory architecture. Some failures are
caused by tool errors, timeouts, unavailable web pages, evaluator ambiguity, or
reasoning mistakes unrelated to memory. FGMD first applies a rule-based scope gate
to remove such out-of-scope cases. A failure is considered memory-attributable only
if the trace contains evidence that memory could plausibly have changed the outcome,
such as missing reusable information, irrelevant retrieved memory, stale memory, or
conflicting memory entries.

Formally, for each failed trajectory $\tau_i$, the scope gate assigns
\[
g_i
=
\operatorname{Scope}(\tau_i)
\in
\{0,1\},
\]
where $g_i=1$ indicates an in-scope memory-related failure. Only trajectories with
$g_i=1$ are passed to module attribution.

\paragraph{Module attribution.}
FGMD first applies a rule-based classifier that assigns each in-scope failure a
dominant failure mode from memory-trace signals---whether relevant units exist in
the pool, whether they were retrieved, whether a consistency judge kept or dropped
them, and whether kept units were stale:
\[
f_i
=
\operatorname{Classify}(\tau_i)
\in
\mathcal{F}.
\]
Here $\mathcal{F}$ collects memory-failure modes such as extraction
gap, low-quality extraction, retrieval miss, judge-rejected retrieval, retrieval
noise, and stale memory, and a fixed table maps each mode to the responsible module
$b_i\in\{E,S,R,M\}$. A failure maps to Encode ($E$) when useful information from
previous experience was never extracted or was compressed into an unusable form; to
Store ($S$) when the information was encoded but represented in a format that
prevents effective access, indexing, or cross-memory linking; to Retrieve ($R$) when
relevant memories exist but are not retrieved, are ranked below irrelevant entries,
or are poorly injected into the agent context; and to Manage ($M$) when the memory
pool contains stale, duplicated, contradictory, or low-value entries that interfere
with current decision making. Failures that resist rule attribution---retrieval and
the consistency judge both pass yet the answer is still wrong---are further examined
by the diagnosis LLM, which reads up to a fixed number of such traces (sampling when
more exist), inspects each one (its question, gold and agent answers, and final
steps), and relabels recognizable non-memory failures (numeric, format, or
time-window errors) as out of scope, sharpening the per-module histogram.

\paragraph{Aggregation.}
Single-trajectory diagnoses are noisy, so FGMD aggregates the per-task modes across
the round. Mapping each mode $f_i$ to its module yields the per-module failure
histogram
\[
H_t(q)
=
\sum_i \mathbf{1}[g_i=1]\,\mathbf{1}[b_i=q],
\qquad q\in\{E,S,R,M\}.
\]
The histogram nominates a candidate bottleneck, but FGMD fixes the primary
bottleneck $q_t^\star$ with an LLM diagnostician that reads the histogram together
with representative failed trajectories and emits a per-module diagnosis. When the
histogram and the diagnostician disagree, FGMD adopts the diagnostician's module and
records the disagreement. FGMD then samples representative trajectories for
$q_t^\star$ as grounded evidence, which prevents a single anomalous failure from
dominating the next search step.

\paragraph{Textual gradient synthesis.}
Finally, an LLM-based synthesizer consolidates the round's diagnostic
signals---the per-module failure histogram, the per-module diagnosis, and the
grounded evidence---into a structured textual gradient:
\[
\delta_{t+1}
=
\left(
q_t^\star,\ 
\mathcal{E}_t,\ 
\rho_t,\ 
\mathcal{R}_t
\right),
\]
where $q_t^\star$ is the primary module bottleneck, $\mathcal{E}_t$ is the grounded
evidence (representative failed-task identifiers), $\rho_t$ is a categorical
confidence level (high, medium, or low), and $\mathcal{R}_t$ is a recommended
architecture edit constrained to the valid edits of the search space. For example, a
Retrieve bottleneck may recommend switching from pure semantic retrieval to hybrid
retrieval or adding diversity-aware reranking, whereas a Manage bottleneck may
recommend deduplication, conflict resolution, or more conservative forgetting. The
synthesized object also carries an out-of-scope flag that realizes the scope
filtering above and a cross-source-agreement field that records whether the
rule-based histogram and the LLM diagnosis agree on $q_t^\star$.
Appendix~\ref{app:worked-delta} shows a complete synthesized gradient produced during
search.

\paragraph{Differential validation.}
To avoid repeatedly following an unhelpful diagnostic direction, FGMD compares the
current round with the previous round. If the last edit targeted module $q$ but did
not improve accuracy, memory-induced gain, or validation performance, the
corresponding recommendation is downgraded in $\mathcal{L}$. If it improves the
Pareto front or validation leader, the recommendation is promoted as an active
principle. This closes the loop between failure attribution and architecture search:
diagnosis proposes a module-level direction, EGAS tests it through candidate
architectures, and validation determines whether the direction should be retained or
refuted.

\section{Experiments}
\label{sec:experiments}

\subsection{Experimental Setup}
\label{sec:exp-setup}

\textbf{Datasets and Metrics.}
We evaluate our method on three challenging agentic search and reasoning benchmarks: GAIA, WebWalkerQA, and xBench-DeepSearch. GAIA provides three difficulty levels, L1/L2/L3, and is therefore used for difficulty-stratified analysis \cite{mialon2023gaia}. WebWalkerQA evaluates deep web navigation and multi-step information seeking \cite{wu2025webwalker}. xBench-DeepSearch contains 100 Chinese deep-search questions and serves as a hard external robustness check \cite{chen2025xbench}. Since xBench-DeepSearch does not provide difficulty labels, all difficulty-based analysis is conducted on GAIA. We evaluate performance using accuracy, memory lift, token cost, latency, number of reasoning steps, retrieval hit-rate, and regression rate, following the definitions in \S\ref{sec:prelim-formulation}. 

\textbf{Baselines.}
We compare our method with a No-Memory baseline and a representative set of memory-based agent systems. These baselines cover major paradigms in agent memory research, including episodic and long-term memory methods, such as Generative Agents \cite{park2023generative}, Mem0 \cite{chhikara2025mem0}, and MemoryBank \cite{zhong2023memorybank}; experiential and reflective memory methods, represented by ExpeL \cite{zhao2024expel} and DiLu \cite{wen2024dilu}; procedural and skill memory methods, including Voyager \cite{wang2023voyager}, Agent Workflow Memory \cite{wang2024awm}, Agent-KB \cite{tang2025agentkb}, Dynamic Cheatsheet \cite{suzgun2025dynamiccheatsheet}, and Memp \cite{fang2025memp}; as well as self-evolving memory systems MemEvolve \cite{zhang2025memevolve} and Mobile-Agent-E \cite{wang2025mobileagente}. This comparison provides broad coverage of existing memory designs for LLM agents.

\textbf{Implementation Details.}
For the task agent, we use Qwen3.5-122B-A10B and gpt-5.1-mini as backbone models, implemented on top of a DAG-based parallel web-agent framework \cite{qin2025flashsearcher}. To avoid overfitting during memory architecture search, we split the data into three disjoint parts: the search split is used to evaluate candidate memory architectures and update $\graph$ and $\ledger$; the held-out validation split is used to select the Pareto leader and perform round-over-round improvement checks; and the test split is used only for final reported results. In each search round, the Meta-LLM proposes $K{=}3$ candidate memory architectures, and the search runs for a small number of rounds per benchmark; the full search budgets are reported in Appendix Table~\ref{tab:search-budget}.

\begin{table}[t]
  \centering
  \caption{Main results: \textsc{AutoMem} vs.\ existing memory architectures on the shared
  benchmark suite. \textbf{Perf.}\ is accuracy
  (\%, $\uparrow$); \textbf{Cost} is mean tokens/task in thousands (k),
  \textbf{Delay} (s), and \textbf{\#Steps} are $\downarrow$ (mean values).
  Rows are grouped by task-agent backbone; within each group the final row is the
  architecture \textsc{AutoMem} discovers (its accuracy shown in \textbf{bold}).}
  \label{tab:main-results}
  {\footnotesize\setlength{\tabcolsep}{2pt}%
  \begin{tabular}{l cccc cccc cccc}
    \toprule
    \multirow{2}{*}{Memory Setting} & \multicolumn{4}{c}{GAIA} & \multicolumn{4}{c}{xBench} & \multicolumn{4}{c}{WebWalkerQA} \\
    \cmidrule(lr){2-5}\cmidrule(lr){6-9}\cmidrule(lr){10-13}
    & Perf. & Cost & Delay & \#Steps & Perf. & Cost & Delay & \#Steps & Perf. & Cost & Delay & \#Steps \\
    \midrule
    \multicolumn{13}{l}{\textit{Backbone: Qwen3.5-122B-A10B}} \\
    No-Memory                                              & 65.0 & 335.5 & 661 & 13.1 & 30.0 & 412.6 & 433 & 14.0 & 67.6 & 111.7 & 309 & 7.3 \\
    Mem0 \cite{chhikara2025mem0}                          & 67.1 & 215.7 & 365 & 9.7 & 36.0 & 434.4 & 621 & 14.3 & 66.5 & 131.1 & 383 & 7.0 \\
    MemoryBank \cite{zhong2023memorybank}                 & 67.8 & 200.4 & 302 & 9.5 & 45.0 & 430.2 & 535 & 14.9 & 67.1 & 111.8 & 283 & 6.6 \\
    ExpeL \cite{zhao2024expel}                            & 63.6 & 204.0 & 329 & 9.7 & 44.0 & 351.7 & 395 & 12.9 & 67.1 & 114.2 & 226 & 7.1 \\
    Voyager \cite{wang2023voyager}                        & 66.4 & 235.7 & 358 & 11.0 & 40.0 & 383.6 & 425 & 13.8 & 66.9 & 130.0 & 260 & 8.0 \\
    Agent-KB \cite{tang2025agentkb}                       & 64.3 & 221.6 & 422 & 10.0 & 40.0 & 338.8 & 458 & 12.4 & 68.9 & 119.6 & 304 & 6.9 \\
    Memp \cite{fang2025memp}                              & 66.4 & 199.1 & 316 & 9.8 & 41.0 & 332.4 & 345 & 12.5 & 66.5 & 104.5 & 215 & 6.7 \\
    \textsc{AutoMem} (ours)                               & \textbf{71.5} & 128.4 & 270 & 6.0 & \textbf{46.0} & 395.7 & 729 & 15.6 & \textbf{72.5} & 118.7 & 182 & 6.4 \\
    \midrule
    \multicolumn{13}{l}{\textit{Backbone: gpt-5.1-mini}} \\
    No-Memory                                             & 69.1 & 86.0 & 505 & 10.4 & 69.0 & 141.0 & 523 & 14.7 & 71.2 & 48.0 & 252 & 6.9 \\
    Generative Agents \cite{park2023generative}           & 66.7 & 61.0 & 436 & 8.9 & 70.0 & 131.0 & 818 & 13.5 & 72.4 & 45.0 & 269 & 6.6 \\
    Voyager \cite{wang2023voyager}                        & 69.7 & 60.0 & 500 & 9.3 & 68.0 & 117.0 & 553 & 12.7 & 73.5 & 49.0 & 334 & 7.0 \\
    DiLu \cite{wen2024dilu}                               & 66.7 & 59.0 & 445 & 8.9 & 69.0 & 134.0 & 501 & 13.8 & 72.9 & 46.0 & 272 & 7.0 \\
    ExpeL \cite{zhao2024expel}                            & 66.1 & 59.0 & 500 & 8.7 & 64.0 & 123.0 & 710 & 13.1 & 69.4 & 76.0 & 385 & 11.0 \\
    Agent Workflow Memory \cite{wang2024awm}              & 67.3 & 62.0 & 585 & 10.2 & 71.0 & 138.0 & 761 & 14.1 & 72.4 & 68.0 & 397 & 11.4 \\
    Mobile-Agent-E \cite{wang2025mobileagente}            & 69.1 & 65.0 & 322 & 9.4 & 68.0 & 120.0 & 537 & 13.2 & 71.8 & 59.0 & 296 & 6.5 \\
    Dynamic Cheatsheet \cite{suzgun2025dynamiccheatsheet} & 68.5 & 69.0 & 560 & 9.7 & 65.0 & 174.0 & 818 & 16.0 & 72.9 & 57.0 & 367 & 7.6 \\
    MemEvolve \cite{zhang2025memevolve}                   & 73.3 & 85.0 & 693 & 10.1 & 74.0 & 136.0 & 773 & 14.2 & 74.7 & 40.0 & 332 & 6.6 \\
    \textsc{AutoMem} (ours)                               & \textbf{75.2} & 103.4 & 832 & 13.1 & \textbf{79.0} & 243.4 & 872 & 11.3 & \textbf{76.5} & 76.6 & 750 & 10.8 \\
    \midrule
    \multicolumn{13}{l}{\textit{Backbone: gpt-5.4}} \\
    \textsc{AutoMem} (ours)                               & \textbf{77.6} & 98.1 & 606 & 11.2 & \textbf{84.0} & 116 & 402 & 8.3 & \textbf{79.4} & 87.9 & 151 & 7.4 \\
    \bottomrule
  \end{tabular}%
  }
\end{table}

\subsection{Main Results}
\label{sec:exp-main}

Table~\ref{tab:main-results} compares \textsc{AutoMem} with representative manually designed memory mechanisms under two task-agent backbones. To avoid conflating memory design with backbone capacity, we compare methods within each backbone group. Under Qwen3.5-122B-A10B, the architecture discovered by \textsc{AutoMem} achieves the best accuracy on all three benchmarks, reaching $71.5\%$ on GAIA, $46.0\%$ on xBench, and $72.5\%$ on WebWalkerQA. Compared with the strongest manually designed baseline on each benchmark, \textsc{AutoMem} improves accuracy by $+3.7$, $+1.0$, and $+3.6$ points, respectively. The same trend holds under gpt-5.1-mini, where \textsc{AutoMem} further improves over the strongest baseline by $+1.9$, $+5.0$, and $+1.8$ points on the three benchmarks. These consistent gains across both backbones show that \textsc{AutoMem} discovers task-adaptive memory architectures that are stronger than fixed human-designed memory mechanisms.

The gains are not simply obtained by blindly increasing memory usage or interaction length. Under Qwen3.5-122B-A10B, \textsc{AutoMem} substantially reduces token cost, delay, and average steps on GAIA, and achieves the lowest delay and fewest steps on WebWalkerQA while maintaining the best accuracy. On xBench, it also improves accuracy over all baselines while using fewer tokens than the closest accuracy competitor, MemoryBank, although the deeper search process leads to higher delay and more steps. In contrast, manually designed memory mechanisms are highly task-sensitive: MemoryBank performs strongly on xBench but brings limited gains on WebWalkerQA, while Agent-KB performs well on WebWalkerQA but falls below the No-Memory baseline on GAIA. These results suggest that different tasks require different memory designs, and that searching task-adaptive Encode/Store/Retrieve/Manage architectures provides a more robust accuracy-efficiency trade-off than relying on a single hand-crafted memory mechanism.




\begin{figure}[t!]
  \centering
  \includegraphics[width=0.98\linewidth]{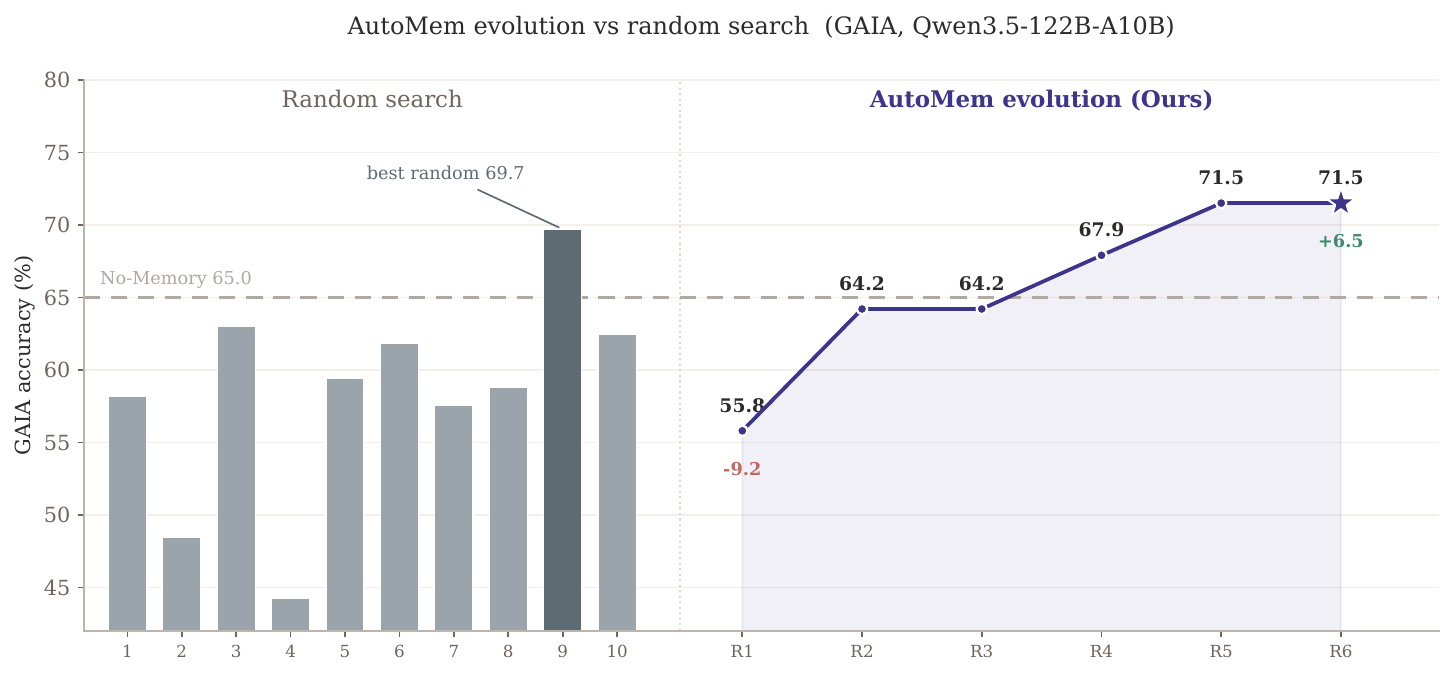}
  \caption{\textsc{AutoMem} evolution vs.\ random search on GAIA
  (Qwen3.5-122B-A10B). \textbf{Left:} ten architectures sampled by random search
  (gray bars), most below the No-Memory baseline (dashed, 65.0) and the best
  reaching 69.7. \textbf{Right:} \textsc{AutoMem}'s best-so-far accuracy over
  rounds, improving monotonically to 71.5 and surpassing the best random sample
  within fewer evaluations.}
  \label{fig:random-search-comparison}
\end{figure}

\subsection{Guided Search vs.\ Random Search}
\label{sec:exp-random}

A natural question is whether the failure-guided evolution in \textsc{AutoMem} is genuinely more effective than blindly sampling architectures from the same factored space. Figure~\ref{fig:random-search-comparison} compares \textsc{AutoMem} with random search under the same evaluation setting. The results show three clear observations. First, \textsc{AutoMem} is substantially more search-efficient: after only five evolution rounds, it reaches $71.5\%$ accuracy, already surpassing the best result obtained by ten random trials ($69.7\%$). This indicates that \textsc{AutoMem} can identify stronger memory architectures with roughly half the search budget. Second, the performance of \textsc{AutoMem} improves steadily across iterations, increasing from $55.8\%$ in the first round to $64.2\%$, $67.9\%$, and finally $71.5\%$. This monotonic improvement suggests that the accumulated reflections and failure-guided feedback provide useful directional signals for architecture refinement. Third, random search is highly unstable: its performance fluctuates substantially across trials, and most sampled architectures even perform worse than the No-Memory baseline ($65.0\%$). These results demonstrate that the gains of \textsc{AutoMem} come from directed, feedback-driven search rather than from simply sampling more candidate architectures.

\paragraph{Token cost.}
Table~\ref{tab:search-cost} complements Figure~\ref{fig:random-search-comparison}
by reporting what each search protocol costs in task-agent rollout tokens. One
full \textsc{AutoMem} evolution run---including the no-memory baseline
evaluation, warm-up, and held-out validation---consumes $142.7$M tokens on GAIA
and $160.1$M on xBench-DeepSearch, whereas scoring the ten sampled architectures
on the full benchmarks costs $259.0$M and $339.0$M: \textsc{AutoMem} completes
its entire directed search for $0.55\times$ and $0.47\times$ the token budget of
random search. Cheaper small-batch scoring does not rescue random search: its
batch-selected champion overfits the search batch and drops to $63.0\%$ on the
full GAIA set, below the No-Memory baseline (Table~\ref{tab:ablation}), so
reliable selection forces random search to re-evaluate every sample at full
scale. \textsc{AutoMem} instead converts each failed rollout into reusable
feedback, reaching a stronger architecture ($71.5$ vs.\ $69.7$) at roughly half
the token cost.

\begin{table}[t!]
  \centering
  \caption{Search-stage token cost of \textsc{AutoMem} vs.\ random search
  (task-agent rollout tokens, in millions; Qwen3.5-122B-A10B backbone).
  \textsc{AutoMem} covers one full evolution run (no-memory baseline, warm-up,
  all search rounds, and held-out validation); random search covers evaluating
  the ten architectures sampled in \S\ref{sec:preliminary} on the full
  benchmark.
  Meta-LLM proposal and diagnosis calls are excluded (a few calls per round,
  negligible relative to rollouts). The random-search totals reuse cached
  search-batch rollouts where available, and one xBench architecture terminated
  early (56/100 tasks), so they are lower bounds.}
  \label{tab:search-cost}
  \small\setlength{\tabcolsep}{6pt}
  \begin{tabular}{l l c c}
    \toprule
    Benchmark & Search protocol & Tokens (M) $\downarrow$ & Rel.\ cost $\downarrow$ \\
    \midrule
    \multirow{2}{*}{GAIA}
      & Random search  & 259.0 & $1.00\times$ \\
      & \textsc{AutoMem} (one evolution run)            & \textbf{142.7} & $\mathbf{0.55\times}$ \\
    \midrule
    \multirow{2}{*}{xBench-DS}
      & Random search  & 339.0 & $1.00\times$ \\
      & \textsc{AutoMem} (one evolution run)            & \textbf{160.1} & $\mathbf{0.47\times}$ \\
    \bottomrule
  \end{tabular}
\end{table}


\begin{table}[t!]
  \centering
  \caption{Component ablation on GAIA.}
  \label{tab:ablation}
  \setlength{\tabcolsep}{6pt}
  \begin{tabular}{l ccccc}
    \toprule
    Variant & Acc.\ $\uparrow$ & Mem.\ lift $\uparrow$ & Hit-rate $\uparrow$ & Tok./task (k) $\downarrow$ & Rounds $\downarrow$ \\
    \midrule
    \textsc{AutoMem} (full: FGMD $+$ EGAS)        & \textbf{71.5} & $\mathbf{+6.5}$ & 81.2 & \textbf{128.4} & 5 \\
    \quad w/o FGMD (generic failure signal)       & 64.2 & $-0.8$  & 78.8 & 226.7 & 5 \\
    \quad w/o EGAS (memoryless proposer)          & 57.9 & $-7.1$  & \textbf{81.7} & 297.3 & 5 \\
    \quad w/o both ($=$ random search)            & 63.0 & $-2.0$ & 6.1 & 193.1 & 5 \\
    \bottomrule
  \end{tabular}
\end{table}


\subsection{Ablation Study}
\label{sec:exp-ablation}

\textsc{AutoMem} consists of two components: Failure-Guided Module Diagnosis (FGMD), which turns failed rollouts into module-level textual feedback, and Experience-Guided Architecture Search (EGAS), which proposes candidates from accumulated search experience. Table~\ref{tab:ablation} ablates both components on GAIA. Removing FGMD replaces structured module diagnosis with a generic ``some tasks failed'' signal while preserving the experienced proposer. Removing EGAS makes the proposer memoryless, allowing it to see only the current architecture and latest diagnosis, without the ledger, reflections, or accumulated search history. Removing both reduces the method to random search.

The results show that both components are necessary. Without FGMD, the discovered architecture drops to $64.2\%$ accuracy with a $-0.8$ memory lift, falling below the No-Memory baseline, while token cost nearly doubles from $128.4$k to $226.7$k. This suggests that, without module-level attribution, the search is easily steered toward costly but ineffective architectures. Removing EGAS is even more damaging, reducing accuracy to $57.9\%$ with a $-7.1$ memory lift and the highest token cost among all variants ($297.3$k). Without the ledger and observation graph, the proposer cannot accumulate reliable search experience: its per-round best score on the search batch decays from $0.70$ to $0.68$, $0.66$, and $0.66$, later rounds re-propose near-identical configurations, and harmful design choices are repeatedly explored. Moreover, the round-1 champion overfits the 50-task search batch, scoring $0.70$ there but falling below the No-Memory baseline on the full set. These two failure modes are complementary: w/o FGMD lacks directional failure attribution, whereas w/o EGAS cannot accumulate and trust useful directions. This supports the text-gradient view in \S\ref{sec:method}, where FGMD provides the gradient signal and EGAS provides the update memory. Finally, removing both yields only $63.0\%$ accuracy with a $-2.0$ lift after four random sampling rounds, confirming that the gains of \textsc{AutoMem} come from directed search rather than from drawing more candidates.

\begin{figure}[t!]
  \centering
  \includegraphics[width=\linewidth]{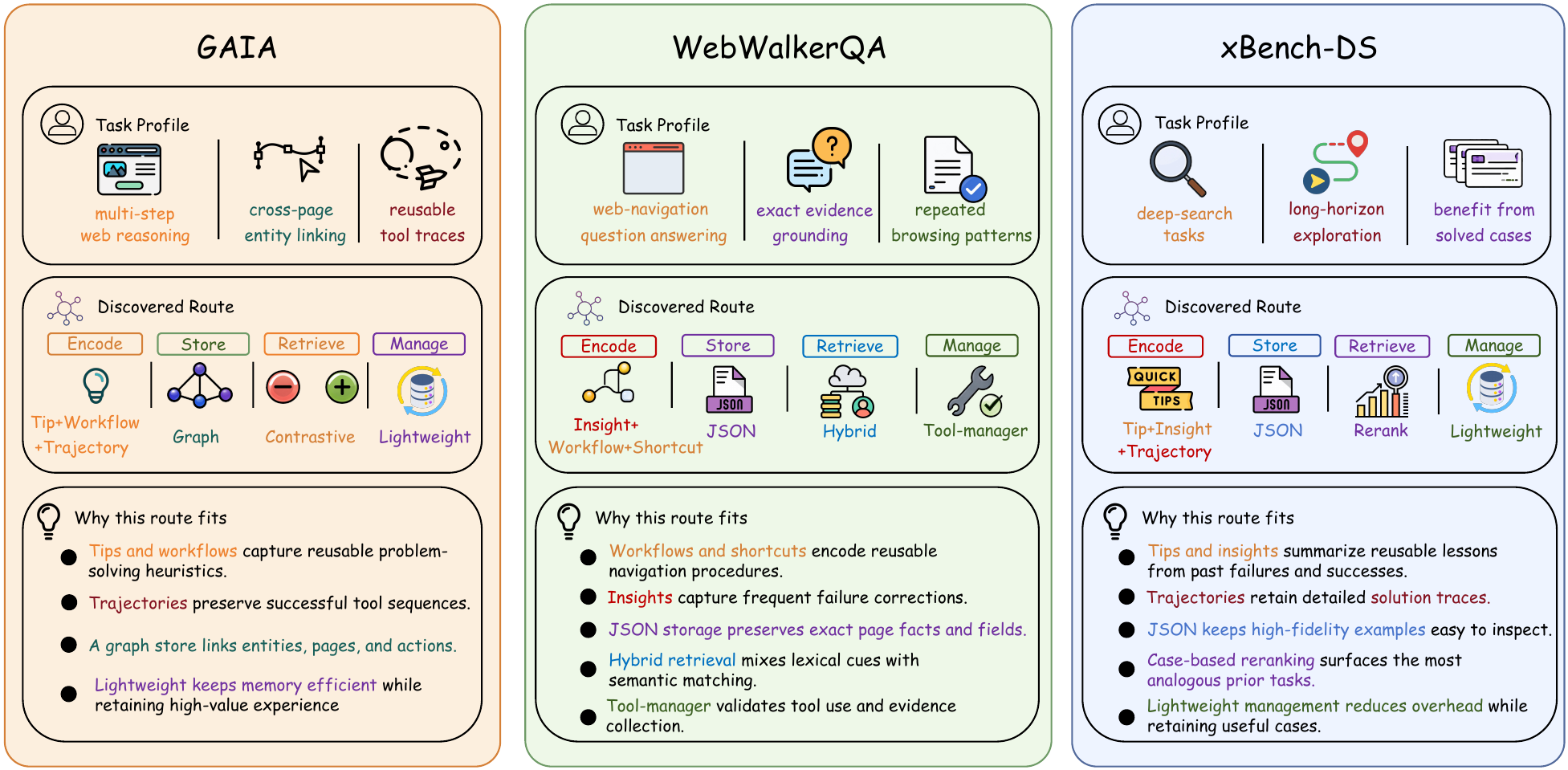}
  \caption{Architectures discovered by \textsc{AutoMem} per benchmark. For each
  benchmark we show its task profile, the discovered Encode/Store/Retrieve/Manage
  route, and why that route fits the task distribution. 
  }
  \label{fig:discovered-architectures}
\end{figure}

\subsection{Discovered Architectures}
\label{sec:exp-arch}

Finally, Figure~\ref{fig:discovered-architectures} visualizes the final Encode/Store/Retrieve/Manage paths discovered by \textsc{AutoMem} on each benchmark, together with the corresponding task profile and the rationale behind each route. Although all architectures are searched from the same factored component menu, \textsc{AutoMem} converges to clearly different memory designs across task distributions. On GAIA, where tasks involve multi-step web reasoning, cross-page entity linking, and reusable tool traces, \textsc{AutoMem} selects \emph{Tip+Workflow+Trajectory} encoding, a graph store, contrastive retrieval, and lightweight management. This route preserves successful tool-use trajectories, organizes entities and actions relationally, and uses contrastive retrieval to distinguish useful experiences from failed ones. On WebWalkerQA, which emphasizes web-navigation question answering, exact evidence grounding, and repeated browsing patterns, the discovered route uses \emph{Insight+Workflow+Shortcut} encoding, a \texttt{JSON} store, hybrid retrieval, and tool-manager based management. This design captures reusable navigation procedures, keeps exact page-level facts inspectable, and validates tool use during evidence collection. On xBench-DS, which requires deep search, long-horizon exploration, and reuse of solved cases, \textsc{AutoMem} selects \emph{Tip+Insight+Trajectory} encoding, a \texttt{JSON} store, case-based reranking, and lightweight management, enabling the agent to reuse prior solution traces while keeping retrieval efficient.

These discovered routes provide qualitative evidence for the central claim of \textsc{AutoMem}: memory architecture should adapt to the task distribution rather than remain fixed. Different benchmarks favor different combinations of what to encode, how to store it, how to retrieve it, and how to manage it. In particular, GAIA benefits from relational graph organization, WebWalkerQA favors structured factual storage and tool validation, while xBench-DS relies more on case reuse and reranking for deep-search reasoning. The diversity of these E/S/R/M paths explains why manually designed memory mechanisms can be highly task-sensitive, and further supports treating long-term memory design as a task-adaptive architecture search problem.



\section{Conclusion}
\label{sec:conclusion}

In this paper, we study long-term memory design for LLM agents as a task-adaptive architecture search problem. We factorize agent memory into Encode, Store, Retrieve, and Manage modules, and show that fixed human-designed memory architectures are highly sensitive to task distributions and backbone models, with no single design consistently dominating across benchmarks. To address this challenge, we propose \textsc{AutoMem}, a text-gradient recursive self-improvement framework that combines Experience-Guided Architecture Search with Failure-Guided Module Diagnosis. By converting failed rollouts into module-level textual feedback and accumulating search experience across iterations, \textsc{AutoMem} efficiently discovers stronger memory architectures under limited evaluation budgets. Experiments on GAIA, WebWalkerQA, and xBench-DeepSearch demonstrate that the discovered architectures outperform existing memory baselines while improving the accuracy-efficiency trade-off, and ablation studies confirm that both failure-guided diagnosis and experience-guided search are essential. Overall, our results suggest that memory for LLM agents should not be treated as a fixed hand-crafted mechanism, but as an adaptive architecture that can be optimized according to the target task distribution.

\section{Limitations and Future Work}
\label{sec:limitation}

This work takes a first step toward task-adaptive memory architecture search, but several directions remain open for future exploration. First, \textsc{AutoMem} currently searches memory architectures at the task or benchmark level, where one discovered Encode/Store/Retrieve/Manage path is used for all samples from the same task distribution. This design is effective and efficient when tasks share similar memory requirements, but individual samples may still differ in their need for memory granularity, retrieval depth, or management strategy. A promising direction is to extend \textsc{AutoMem} from task-level architecture search to sample-level or episode-level memory adaptation, where the agent dynamically selects or routes memory modules according to the current query, interaction history, and observed failure signals.

Second, the current search space is constructed from memory components inspired by existing memory frameworks. This makes the search process interpretable, controllable, and comparable with prior human-designed systems, but it also means that \textsc{AutoMem} mainly discovers new combinations of known memory mechanisms. Future work can move beyond recombination and explore open-ended memory architecture generation, where the system can propose entirely new memory modules, interfaces, update rules, or cross-module interaction patterns. Coupling such architecture invention with executable validation and cost-aware evaluation may further enable LLM agents to evolve from selecting among predefined memory designs toward creating genuinely novel and self-improving memory systems.


\bibliographystyle{unsrt}
\bibliography{ref}

\appendix
\section{Experience Extraction Prompts}
\label{app:prompts}
\begin{promptbox}[Memory Extraction Prompt (placeholder)]
You are extracting reusable memory from an agent trajectory.
Given the observation graph G and the task, decide which memory types to extract
(tip / insight / trajectory / workflow / shortcut) and emit structured MemoryUnits.
...
\end{promptbox}
Example units produced by these prompts are shown in
Appendix~\ref{app:worked-units}.

\section{Worked Examples from the Search State}
\label{app:worked}
This appendix shows what the optimization state of \textsc{AutoMem} looks like
at runtime. All artifacts are verbatim outputs from search runs on GAIA,
WebWalkerQA, and xBench-DeepSearch, lightly abridged for presentation: ellipses
mark omitted fields, floating-point digits are truncated, and non-ASCII symbols
are rendered in ASCII (e.g.\ \texttt{->}). Accuracies appearing inside these
artifacts are intermediate statistics on small search batches; they illustrate
the mechanism and are not the results reported in \S\ref{sec:experiments}.

\subsection{One Optimization Round, End to End}
\label{app:worked-round}
We trace rounds 2--3 of a search run on xBench-DeepSearch. At round~2 the
incumbent architecture extracts tip, trajectory, and insight units into
all-JSON storage with hybrid retrieval and lightweight management. FGMD
aggregates the round's failures into the textual gradient below. Note the
scope gate at work: the \texttt{budget\_capped} failures are tagged
out-of-scope and do not drive architecture edits.

\begin{examplebox}[Round-2 textual gradient (xBench-DeepSearch run, abridged)]
{"total_tasks": 30, "success_count": 13, "failure_count": 17,
 "breakdown": {"retrieval_noise": 9, "budget_capped": 6,
               "extraction_gap": 2, ...},
 "layer_diagnosis": {
  "encode": {"status": "warning",
   "observation": "Encode inserted 63 units (avg 2.1 per task), but 2
     extraction_gap failures had 0 relevant units, e.g. tasks 122 and 156.",
   ...},
  "store": {"status": "ok", ...},
  "retrieve": {"status": "warning",
   "observation": "Retrieve is noisy: 9 retrieval_noise failures; tasks 136,
     149, and 153 had judge-passed units despite TF-IDF finding 0 relevant
     memories.",
   "suggestion": "Switch retrieval strategy to a more relevance-discriminative
     option to reduce off-topic candidates."},
  "manage": {"status": "ok", ...},
  "overall": "Primary bottleneck is Retrieve: retrieval_noise accounts for 9
    failures, while accuracy only rose slightly from 0.40 to 0.433.",
  "priority_action": "retrieval: hybrid -> contrastive"}}
\end{examplebox}

The gradient names one coordinate (Retrieve) and a direction. In round~3 the
proposer emits single-coordinate edits of the incumbent along this gradient:
one candidate applies the suggested edit (retrieval
\texttt{hybrid -> contrastive}), and a second probes an alternative option on
the same coordinate (\texttt{hybrid -> cbr\_rerank}), keeping every other
module fixed so that the round yields a controlled comparison.

\begin{examplebox}[Round-3 candidate proposal (verbatim)]
{"candidate_id": 1,
 "hypothesis": "Replace the champion's hybrid retriever with cbr_rerank to
   reduce dominant retrieval_noise by applying case-based candidate selection
   plus LLM reranking before memory injection.",
 "architecture": {
  "extract_types": ["tip", "trajectory", "insight"],
  "storage_routing": {"tip": "json", "trajectory": "json", "insight": "json"},
  "retrieval": "cbr_rerank",
  "management": "lightweight"},
 "diversity_role": "exploit_a",
 "rationale": "Inherited from the champion: extract_types, storage_routing,
   and management; mutated only retrieval from hybrid to cbr_rerank. The
   dominant attribution category is retrieval_noise: r2_c0 had 9
   retrieval_noise failures, with tasks such as 136, 149, and 153 receiving
   judge-passed but irrelevant memories. Pool counts for tip, trajectory, and
   insight are all above json's threshold, so this keeps the stable
   low-cold-start storage path while testing an under-sampled retrieval option
   designed to improve relevance discrimination rather than merely increase
   hit rate."}
\end{examplebox}

The rationale cites the gradient's evidence (the same failing tasks), the pool
statistics, and explicitly separates inherited from mutated coordinates. In
the paired round-3 evaluation both edits improve on the incumbent (accuracy
0.433 -> 0.467 for contrastive, 0.433 -> 0.500 for cbr-rerank); the cbr-rerank
candidate becomes the new front leader, and the ledger distills the round into
a new principle and a follow-up experiment:

\begin{examplebox}[Ledger update after round 3 (abridged)]
{"id": "P003", "domain": "retrieval", "status": "active", "confidence": 0.4,
 "claim": "With extract_types=tip+trajectory+insight, all-json storage, and
   lightweight management held constant, cbr_rerank beat contrastive on
   hit_rate, accuracy, memory_lift, and fitness in round 3.",
 "evidence": [
  {"round": 3, "candidate": "r3_c1", "metric": "accuracy", "delta": 0.033,
   "supports": true,
   "context": "cbr_rerank r3_c1 accuracy 0.5000 vs contrastive r3_c0 0.4667;
     controlled retrieval-only change"},
  ...]}

{"id": "Q006", "status": "untested", "raised_round": 3,
 "question": "Does cbr_rerank outperform hybrid under the current
   tip+trajectory+insight, all-json, lightweight setup?",
 "recommended_test": "Run a paired candidate identical to r3_c1 except
   retrieval=hybrid, comparing against r3_c1 on accuracy, hit_rate,
   retrieval_noise, and fitness."}
\end{examplebox}

\subsection{A Synthesized Textual Gradient}
\label{app:worked-delta}
The object below is the textual gradient $\delta_{t+1}$ of
\S\ref{sec:method-fgmd} exactly as synthesized at the end of a round (here,
round~1 of a WebWalkerQA run). The fields realize the tuple
$(q_t^\star,\mathcal{E}_t,\rho_t,\mathcal{R}_t)$:
\texttt{primary\_signal} states the dominant bottleneck $q_t^\star$,
\texttt{evidence\_task\_ids} grounds the evidence $\mathcal{E}_t$,
\texttt{confidence} is $\rho_t$, and \texttt{recommended\_action} carries
$\mathcal{R}_t$. The \texttt{out\_of\_scope\_for\_memory} flag implements
scope filtering, and \texttt{cross\_source\_agreement} records whether the
rule-based histogram and the LLM diagnostician point in the same direction.

\begin{examplebox}[Synthesized textual gradient (WebWalkerQA run, round 1)]
{"primary_signal": "Retrieval noise is the dominant failure mode (10/15
   failures). The hybrid retriever returns off-topic units that pass the
   judge, causing 24% task failure. The memory pool lacks coverage for
   out-of-domain topics (3 domain_knowledge_gap tasks with zero relevance),
   but the immediate bottleneck is retrieval quality.",
 "confidence": "high",
 "recommended_action": "Switch retrieval from hybrid to dense-only to reduce
   noise from sparse matching when no relevant units exist.",
 "memory_bypass": false,
 "out_of_scope_for_memory": false,
 "evidence_task_ids": ["EHA_2023_meeting", "SIGCHI_EC_meeting",
   "SIGCHI_CARES_Mullers"],
 "cross_source_agreement": "rule + LLM agree: retrieval_noise is primary
   bottleneck",
 "reasoning": "Primary signals: failure breakdown shows retrieval_noise count
   10 (dominant), LLM 4-layer diagnosis flags retrieval warning and suggests
   hybrid->dense, and top-3 evidence confirms judge passes irrelevant units.
   ..."}
\end{examplebox}

\subsection{Experience-Ledger Entries}
\label{app:worked-ledger}
Each ledger entry realizes the tuple $\ell=(c,u,q,z)$ of
\S\ref{sec:method-egas}: the \texttt{claim} states the condition and the edit,
\texttt{evidence} is the supporting record $q$, and \texttt{status} is the
lifecycle state $z$. The first principle below accumulated consistent evidence
over seven rounds of a GAIA run (20 evidence entries; confidence 0.86) and
encodes a measurement insight---a higher retrieval hit-rate does not imply
higher accuracy. The second shows differential validation at work: a paired
comparison supported the claim in round~6, the same paired comparison
contradicted it in round~7, and the principle was downgraded to
\texttt{refuted}, so the proposer stops following that edit.

\begin{examplebox}[An active and a refuted principle (GAIA run, abridged)]
{"id": "P004", "domain": "retrieval", "status": "active", "confidence": 0.86,
 "claim": "Within round-2 contrastive variants at pool=255, higher hit_rate
   did not translate into higher accuracy when failures were dominated by
   reasoning errors.",
 "evidence": [
  {"round": 2, "candidate": "r2_c0", "metric": "accuracy", "delta": -0.04,
   "supports": true,
   "context": "vs r2_c2 despite hit_rate being higher by 0.149"},
  ... 19 more entries spanning rounds 2-8 ...],
 "first_introduced_round": 2, "last_validated_round": 8}

{"id": "P010", "domain": "general", "status": "refuted", "confidence": 0.25,
 "claim": "At pool=255 with shortcut=json and raw_topk, the contrastive +
   lightweight combo beat the hybrid + tool_manager combo on accuracy and
   fitness despite lower hit_rate.",
 "evidence": [
  {"round": 6, "candidate": "r6_c3", "metric": "accuracy", "delta": 0.08,
   "supports": true,
   "context": "vs r6_c2; same shortcut=json storage and raw_topk"},
  {"round": 7, "candidate": "r7_c3", "metric": "accuracy", "delta": -0.10,
   "supports": false,
   "context": "vs r7_c2; same paired comparison repeated in round 7"},
  ...],
 "first_introduced_round": 6, "last_validated_round": 7}
\end{examplebox}

Dead-end entries record an architecture region that should not be re-proposed
while the entry stays active. The recorded reason matters: in the entry below,
the hit-rate was high and 91\% of the failures were out of memory's
scope, so further retrieval edits in that region would chase noise.

\begin{examplebox}[A dead-end entry (GAIA run, verbatim)]
{"combo": "tip+trajectory+workflow+shortcut extraction with
   tip/trajectory/workflow=hybrid, shortcut=json, hybrid retrieval,
   tool_manager management, and raw_topk injection",
 "outcome": "accuracy=0.36, memory_lift=-0.08, and fitness=0.286 despite
   hit_rate=0.929; 91% of failures were reasoning_error rather than
   retrieval failures",
 "evidence_round": 5, "active": true}
\end{examplebox}

\subsection{An Observation-Graph Excerpt}
\label{app:worked-graph}
The excerpt below shows the observation graph $\mathcal{G}$ of a WebWalkerQA
run after six rounds: two task-pattern nodes and three of its 32 edges. The
edge attributes \texttt{n\_trials} and \texttt{avg\_acc} are exactly the
$n(c,u)$ and the running average behind $\mu(c,u)$ in
\S\ref{sec:method-egas}. The graph is serialized into the proposer prompt as a
soft prior over module choices.

\begin{examplebox}[Observation-graph excerpt (WebWalkerQA run, abridged)]
"nodes": [
 {"id": "Lmedium", "kind": "task_pattern",
  "attrs": {"n_tasks_seen": 28,
            "categories": {"general": 25, "web_research": 3},
            "best_acc_so_far": 0.667, "baseline_acc": 0.75}},
 {"id": "Lcomplex", "kind": "task_pattern",
  "attrs": {"n_tasks_seen": 9, "categories": {"general": 9},
            "best_acc_so_far": 0.667, "baseline_acc": 0.778}}],
"edges": [
 {"src": "Lcomplex", "rel": "arch_evaluated",
  "dst": "extract:insight+tip+trajectory | ret:hybrid | mgmt:lightweight |
          store:insight=json,tip=vector,trajectory=vector",
  "attrs": {"n_trials": 8, "avg_acc": 0.619}},
 {"src": "Lcomplex", "rel": "tried_with",
  "dst": "extract:insight+tip+trajectory",
  "attrs": {"n_trials": 16, "avg_acc": 0.603}},
 {"src": "Lcomplex", "rel": "worked_with_retriever",
  "dst": "retrieval:hybrid",
  "attrs": {"n_trials": 12, "avg_acc": 0.611}}, ...]
\end{examplebox}

\subsection{Example Memory Units}
\label{app:worked-units}
We close with one memory unit per type, extracted by the prompts of
Appendix~\ref{app:prompts} from GAIA trajectories. Tip and insight units are
declarative; trajectory, workflow, and shortcut units are procedural at
increasing levels of abstraction---a grounded trace, a generalized template
with \texttt{<PLACEHOLDER>} slots, and an executable macro with an input
schema. Note that the insight unit is extracted from a \emph{failed} task, and
that every unit carries \texttt{use\_when}/\texttt{avoid\_when} guards that
are checked at injection time.

\begin{examplebox}[Tip unit (declarative heuristic)]
{"type": "tip", "confidence": 1.0, "task_outcome": "success",
 "content": {
  "topic": "Accession pivot",
  "principle": "Do search an accession identifier in the holding
    institution's collection record first, because catalog records provide
    the authoritative subject needed for later biographical hops.",
  "micro_example": "Use the museum catalog page to identify the portrait
    subject before searching ecclesiastical biographical sources.",
  "counterfactual": "Searching the role question first can surface same-name
    or unrelated clergy.",
  "use_when": ["query gives museum accession number",
               "object subject is needed for next hop"],
  "avoid_when": ["question already names the target person"],
  "leakage_risk": "low", "category": "tool_and_search"}}
\end{examplebox}

\begin{examplebox}[Insight unit (root-cause analysis of a failed task)]
{"type": "insight", "confidence": 1.0, "task_outcome": "failure",
 "source_task_query": "Which contributor to the version of OpenCV where
   support was added for the Mask-RCNN model has the same name as a former
   Chinese head of government when the names are transliterated to the Latin
   alphabet?",
 "content": {
  "failure_pattern": "incomplete_search",
  "root_cause_conclusion": "The agent failed to traverse from a feature
    mention to the specific release or commit contributors and then
    cross-check candidate names against an external biographical reference.",
  "corrective_strategy": "For multi-hop entity questions, build a candidate
    table from primary repository evidence, then verify each candidate
    against an independent authority before final_answer.",
  "detection_signal": "Only one source type was consulted for a multi-hop
    identity match.",
  "use_when": ["Question asks for a contributor",
               "Feature must be tied to a release", ...],
  "avoid_when": ["Single authoritative source directly answers"]}}
\end{examplebox}

\begin{examplebox}[Trajectory unit (grounded action--observation trace)]
{"type": "trajectory", "confidence": 1.0, "task_outcome": "success",
 "source_task_query": "How many studio albums were published by Mercedes
   Sosa between 2000 and 2009 (included)? ...",
 "content": {
  "steps": [
   {"action": "web_search(Mercedes Sosa discography Wikipedia)",
    "observation": "Top search result linked to Mercedes Sosa English
      Wikipedia page"},
   {"action": "browser_open(Mercedes Sosa Wikipedia discography section)",
    "observation": "Studio albums table listed 3 releases between 2000 and
      2009"}],
  "key_decision": "Verified counts using the primary entity Wikipedia page
    discography table to ensure release date accuracy",
  "tool_strategy": "web_search -> browser_crawl(Wikipedia_discography)",
  "reusable_anchor": "Wikipedia discography section year-range extraction
    pattern",
  "use_when": ["Query requests item count within date range",
               "Subject has dedicated Wikipedia page"],
  "avoid_when": ["Content behind paywall",
                 "No tabulated discography available"]}}
\end{examplebox}

\begin{examplebox}[Workflow unit (generalized multi-step template)]
{"type": "workflow", "confidence": 1.0, "task_outcome": "success",
 "content": {
  "agent_workflow": [
   {"step": 1,
    "action": "web_search the target runner's major record-setting race
      completion time.",
    "generalized_execution": "Fetch <ENTITY> <METRIC> from primary sports
      archive record."},
   {"step": 2,
    "action": "crawl_page the primary-source encyclopedia infobox containing
      distance metrics.",
    "generalized_execution": "Extract <OBJECT> closest approach distance from
      standard reference infobox."},
   {"step": 3,
    "action": "code_interpreter calculate total duration; scale to target
      units; round integer.",
    "generalized_execution": "Calculate <TOTAL>/<RATE>; divide 1000; round
      to nearest integer."}],
  "chain_type": "web",
  "final_format_check": "Round result to nearest thousand; suppress comma
    separators.", ...}}
\end{examplebox}

\begin{examplebox}[Shortcut unit (executable macro with input schema)]
{"type": "shortcut", "confidence": 1.0, "task_outcome": "success",
 "content": {
  "name": "date_anchored_fact_lookup",
  "description": "Retrieves a specific factual claim attributed to a named
    source category within a defined time period using search-based
    verification.",
  "precondition": "Query requires identifying a value associated with a
    specific source and date that is not in internal knowledge base.",
  "input_schema": [
   {"name": "<SOURCE_CATEGORY>", "type": "string"},
   {"name": "<DATE_RANGE>", "type": "string"},
   {"name": "<FACT_TYPE>", "type": "string"}],
  "action_sequence": [
   {"step": 1, "tool_call": {"tool_name": "web_search",
    "args_pattern": "\"<SOURCE_CATEGORY>\" \"<DATE_RANGE>\" \"<FACT_TYPE>\""}},
   {"step": 2, "tool_call": {"tool_name": "final_answer",
    "args_pattern": "Submit extracted value as answer."}}],
  "assumptions": ["Network access is available.", ...]}}
\end{examplebox}

\section{Implementation Details}
\label{app:impl}
Memory schema, storage backends, retrieval strategies, management ops; the
exact factored search space over $(E,S,R,M)$ and its cross-module constraints;
observation-graph update rule; ledger schema.
Table~\ref{tab:component-lineage} lists the full per-module component menu
together with the prior work each option draws on.

\begin{table}[H]
  \centering
  \caption{The per-module component menu of the factored search space
  (\S\ref{sec:prelim-memory-space}) and its lineage. Each option is selectable and
  routable by the proposer; cross-module constraints (e.g., \texttt{graph}
  retrieval requires a graph-family store) keep every $(E,S,R,M)$ path valid by
  construction.}
  \label{tab:component-lineage}
  \small\setlength{\tabcolsep}{5pt}
  \begin{tabular}{l l l}
    \toprule
    Module & Option & Lineage \\
    \midrule
    \multirow{5}{*}{Encode}
      & \texttt{tip} (transferable heuristics)        & \cite{zhao2024expel, shinn2023reflexion} \\
      & \texttt{insight} (failure root-cause)         & \cite{shinn2023reflexion, zhao2024expel} \\
      & \texttt{trajectory} (action--observation)     & \cite{park2023generative, packer2024memgpt} \\
      & \texttt{workflow} (orchestration logic)       & \cite{wang2024awm, fang2025memp} \\
      & \texttt{shortcut} (parameterized macro)       & \cite{wang2023voyager, zheng2025skillweaver} \\
    \midrule
    \multirow{5}{*}{Store}
      & \texttt{json} (exact key--value)              & --- \\
      & \texttt{vector} (FAISS dense index)           & \cite{chhikara2025mem0, packer2024memgpt} \\
      & \texttt{hybrid} (json $+$ vector)             & --- \\
      & \texttt{graph} (entity--relation, NetworkX)   & \cite{xu2025amem} \\
      & \texttt{llm\_graph} (LLM temporal KG)         & \cite{rasmussen2025zep} \\
    \midrule
    \multirow{6}{*}{Retrieve}
      & \texttt{hybrid} (lexical $+$ semantic)        & --- \\
      & \texttt{contrastive} (success/failure cohort) & \cite{zhao2024expel} \\
      & \texttt{cbr\_rerank} (case-bank $+$ rerank)   & \cite{zhou2025memento} \\
      & \texttt{graph} (multi-hop traversal)          & \cite{xu2025amem} \\
      & \texttt{hyde} (hypothetical-document embed)   & \cite{gao2022hyde} \\
      & \texttt{mmr} (diversity reranking)            & \cite{carbonell1998mmr} \\
    \midrule
    \multirow{4}{*}{Manage}
      & \texttt{lightweight} (\texttt{forget})        & \cite{zhong2023memorybank} \\
      & \texttt{json\_full} (\texttt{merge} $+$ \texttt{update})  & \cite{chhikara2025mem0, wang2024awm} \\
      & \texttt{tool\_manager} (\texttt{validate})    & \cite{wang2023voyager, suzgun2025dynamiccheatsheet} \\
      & \texttt{graph\_consolidate} (\texttt{merge} $+$ \texttt{evolve}) & \cite{xu2025amem, zhou2025memento} \\
    \bottomrule
  \end{tabular}
\end{table}

\section{Algorithm Details}
\label{app:algo}
Pseudocode for the optimization loop and the diagnostic pipeline.

\section{Supplementary Experiments and Reproducibility}
\label{app:supp}

\begin{table}[H]
  \centering
  \caption{Search budget and reproducibility details (cf.\ \S\ref{sec:exp-random}),
  ensuring \textsc{AutoMem}, random search, and the fixed baselines are compared
  under transparent, comparable budgets. All splits are disjoint and stratified
  with a fixed seed (42). On WebWalkerQA and xBench-DS the memory pool is seeded
  by the first search round instead of a separate warm-up split; on WebWalkerQA
  the validation and held-out tasks come from websites unseen during search.}
  \label{tab:search-budget}
  \small
  \begin{tabular}{l ccc}
    \toprule
    & GAIA & WebWalkerQA & xBench-DS \\
    \midrule
    Warm-up (profile) size        & 20 & -- & -- \\
    Search-split size             & 50  & 60  & 50  \\
    Validation size               & 95 & 110 & 50 \\
    Final report set              & 165 (full) & 170 (fixed subset) & 100 (full) \\
    Split stratification          & category $\times$ level & site-clustered & difficulty quadrant \\
    Candidates per round $K$      & 3 & 3 & 3 \\
    Rounds                        & 5 & 6 & 5 \\ 
    Judge model                   & Qwen3.5-122B-A10B & Qwen3.5-122B-A10B & Qwen3.5-122B-A10B \\
    Token budget                  & 8192 & 8192 & 8192 \\
    \bottomrule
  \end{tabular}
\end{table}

\end{document}